\documentclass[10pt,twocolumn,letterpaper]{article}

 \usepackage{cvpr}              % To produce the CAMERA-READY version
\usepackage{comment}

\definecolor{cvprblue}{rgb}{0.21,0.49,0.74}
\usepackage[pagebackref,breaklinks,colorlinks,allcolors=cvprblue]{hyperref}

\usepackage{graphicx}
\usepackage{cuted}
\usepackage{capt-of}

\usepackage{booktabs}  
\usepackage{multirow}   
\usepackage{amssymb}    
\usepackage{amsmath}  
\usepackage{arydshln}
\usepackage{algorithm}
\usepackage{algorithmic}
\usepackage{xcolor}
\usepackage{placeins}

\newcommand{\best}[1]{\textbf{\underline{#1}}}
\newcommand{\secondbest}[1]{\textbf{#1}}

\newcommand{\INPUT}{\item[\textbf{Input:}]}
\newcommand{\OUTPUT}{\item[\textbf{Output:}]}

\def\paperID{17974} % *** Enter the Paper ID here
\def\confName{CVPR}
\def\confYear{2026}

\title{Mesh RAG: Retrieval Augmentation for Autoregressive Mesh Generation}

\author{
Xiatao Sun$^1$\and Chen Liang$^1$\and Qian Wang$^1$\and Daniel Rakita$^1$
\\
\textsuperscript{1}Yale University, New Haven, CT, USA
\\
\textsuperscript{1}\{xiatao.sun, dylan.liang, peter.wang.qw262, daniel.rakita\}@yale.edu
}
\begin{document}
\maketitle

\begin{strip}
  \centering
  \includegraphics[width=\textwidth]{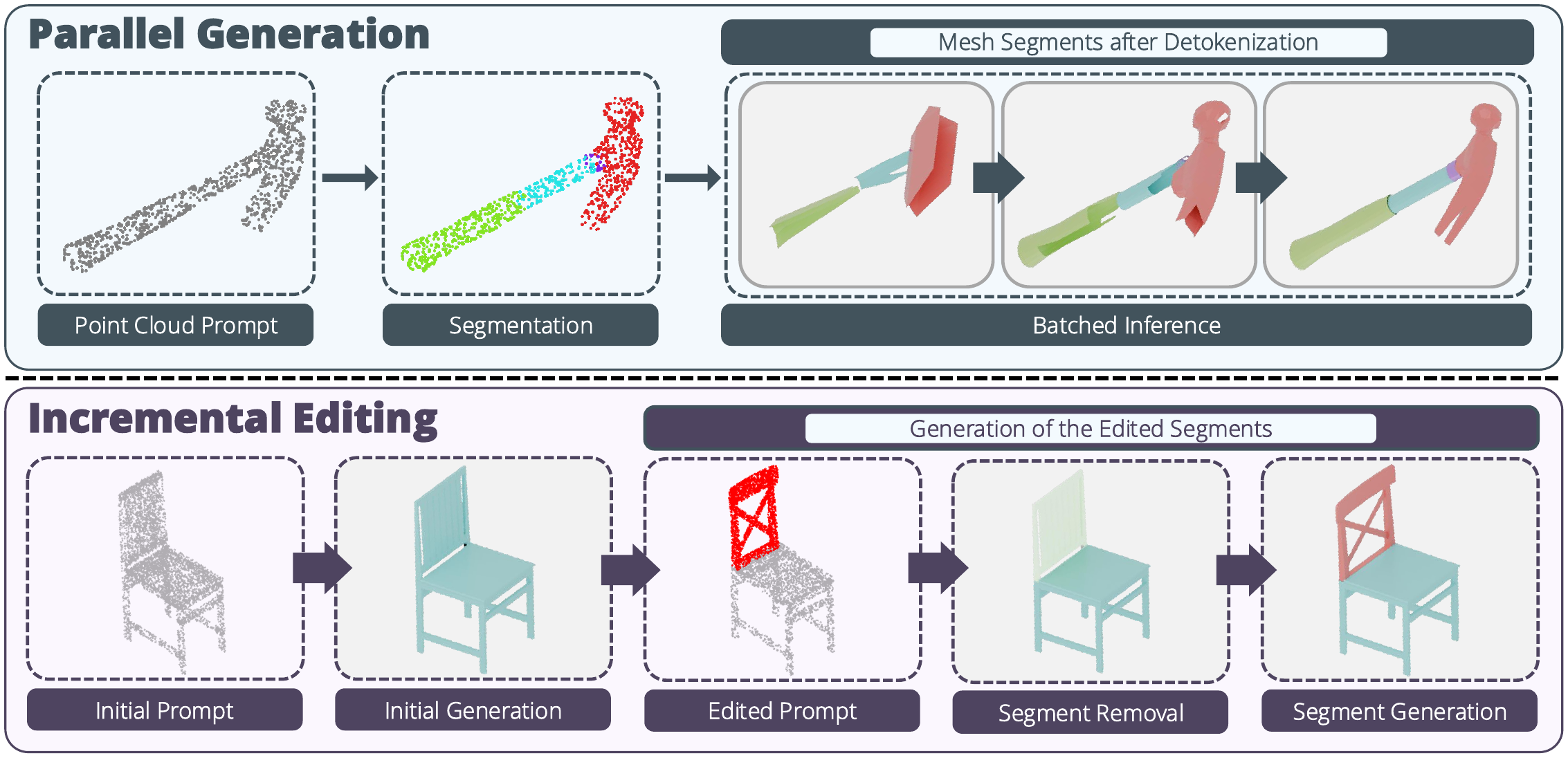}
  \captionof{figure}{\textbf{Parallel generation and incremental editing with Mesh RAG.} Our framework enhances autoregressive models by enabling parallel generation (Top) to improve quality and reduce inference time. It also supports efficient incremental editing (Bottom), allowing for localized generation of newly added or edited segments.} 
  \label{fig:teaser}
\end{strip}

\begin{abstract}

3D meshes are a critical building block for applications ranging from industrial design and gaming to simulation and robotics. Traditionally, meshes are crafted manually by artists, a process that is time-intensive and difficult to scale. To automate and accelerate this asset creation, autoregressive models have emerged as a powerful paradigm for artistic mesh generation. However, current methods to enhance quality typically rely on larger models or longer sequences that result in longer generation time, and their inherent sequential nature imposes a severe quality-speed trade-off. This sequential dependency also significantly complicates incremental editing. To overcome these limitations, we propose Mesh RAG, a novel, training-free, plug-and-play framework for autoregressive mesh generation models. Inspired by RAG for language models, our approach augments the generation process by leveraging point cloud segmentation, spatial transformation, and point cloud registration to retrieve, generate, and integrate mesh components. This retrieval-based approach decouples generation from its strict sequential dependency, facilitating efficient and parallelizable inference. We demonstrate the wide applicability of Mesh RAG across various foundational autoregressive mesh generation models, showing it significantly enhances mesh quality, accelerates generation speed compared to sequential part prediction, and enables incremental editing, all without model retraining.

\end{abstract}    
\section{Introduction}
\label{sec:intro}

Triangular meshes serve as the fundamental geometric representation for a vast array of digital applications, ranging from gaming and visual effects to industrial simulation and robotics. However, producing high-quality, artist-friendly meshes remains a significant bottleneck in production pipelines, as it traditionally relies heavily on time-intensive manual labor by skilled 3D artists. To overcome this scalability challenge, automated mesh generation has become a critical research area. Traditional approaches typically first generate an intermediate representation based on the input prompt and then reconstruct the final mesh \cite{lorensen1998marching, lai2025hunyuan3d}. While capable of producing visually high-fidelity results, these reconstruction-based methods frequently generate dense, irregular topologies that are ill-suited for downstream applications demanding real-time rendering \cite{aktacs2016survey, sun2025prism, sun2025hybrid}.

To address these topological shortcomings, autoregressive models have emerged as a powerful paradigm, drawing inspiration from recent successes in large language models \cite{siddiqui2024meshgpt}. By sequentially generating mesh components as a stream of tokens, these models can produce clean, regular topologies with compact triangular faces reminiscent of those crafted by human artists \cite{chenmeshanything}. However, this sequential dependency introduces a critical bottleneck. Autoregressive mesh generation faces a severe quality-efficiency trade-off, where generating complex meshes requires long inference times or larger models \cite{tangedgerunner, zhao2025deepmesh, hao2024meshtron}. Furthermore, their sequential nature makes parallelization and incremental editing fundamentally challenging.

To overcome these limitations, we introduce Mesh RAG, a novel, training-free, and plug-and-play framework for autoregressive mesh generation. Diverging from most prior efforts that focus on improving tokenization strategies \cite{chen2025meshanything, weng2025scaling, lionar2025treemeshgpt}, our approach is inspired by Retrieval-Augmented Generation (RAG) in language models. Our method decomposes a target shape, provided as a point cloud prompt, into distinct segments. It then generates mesh components for each segment and seamlessly integrates them by transformations and point cloud registration to retrieve their desired positions, orientations and scales. This compositional approach decouples the generation process from its strict sequential dependency, unlocking parallelizable inference and intuitive, localized editing.

We demonstrate the efficacy of Mesh RAG through a set of experiments on several foundational autoregressive models. Our results show that Mesh RAG not only significantly enhances final mesh quality across multiple metrics but also accelerates generation speed compared to predicting each segment sequentially, all without any model retraining. Furthermore, we showcase its unique capability for incremental editing under autoregressive mesh generation.  %We provide open-source implementation of Mesh RAG to facilitate future research on retrieval augmentation on autoregressive mesh generation.\footnote{\textit{[link removed for blind review process]}} 

% In summary, our contributions are:

% and broad applicability 

\begin{comment}
\begin{itemize}
    \item We introduce Mesh RAG, a novel, training-free, and plug-and-play framework that enhances existing autoregressive mesh generation models.
    \item Our framework decouples the inherent sequential dependency of autoregressive models, enabling a parallelizable inference pipeline for better mesh quality and an incremental editing pipeline for enhanced flexibility.
    \item We provide a set of experimental validation showing that Mesh RAG boosts mesh quality and accelerates generation speed compared to sequential segment prediction for various state-of-the-art autoregressive models, establishing a new, more efficient paradigm.
    \item We provide open-source implementation of Mesh RAG to facilitate future research on retrieval augmentation on autoregressive mesh generation.
    \footnote{\textit{[link removed for blind review process]}}
\end{itemize}    
\end{comment}

\begin{figure*}[t]
    \centering
    \includegraphics[width=\textwidth]{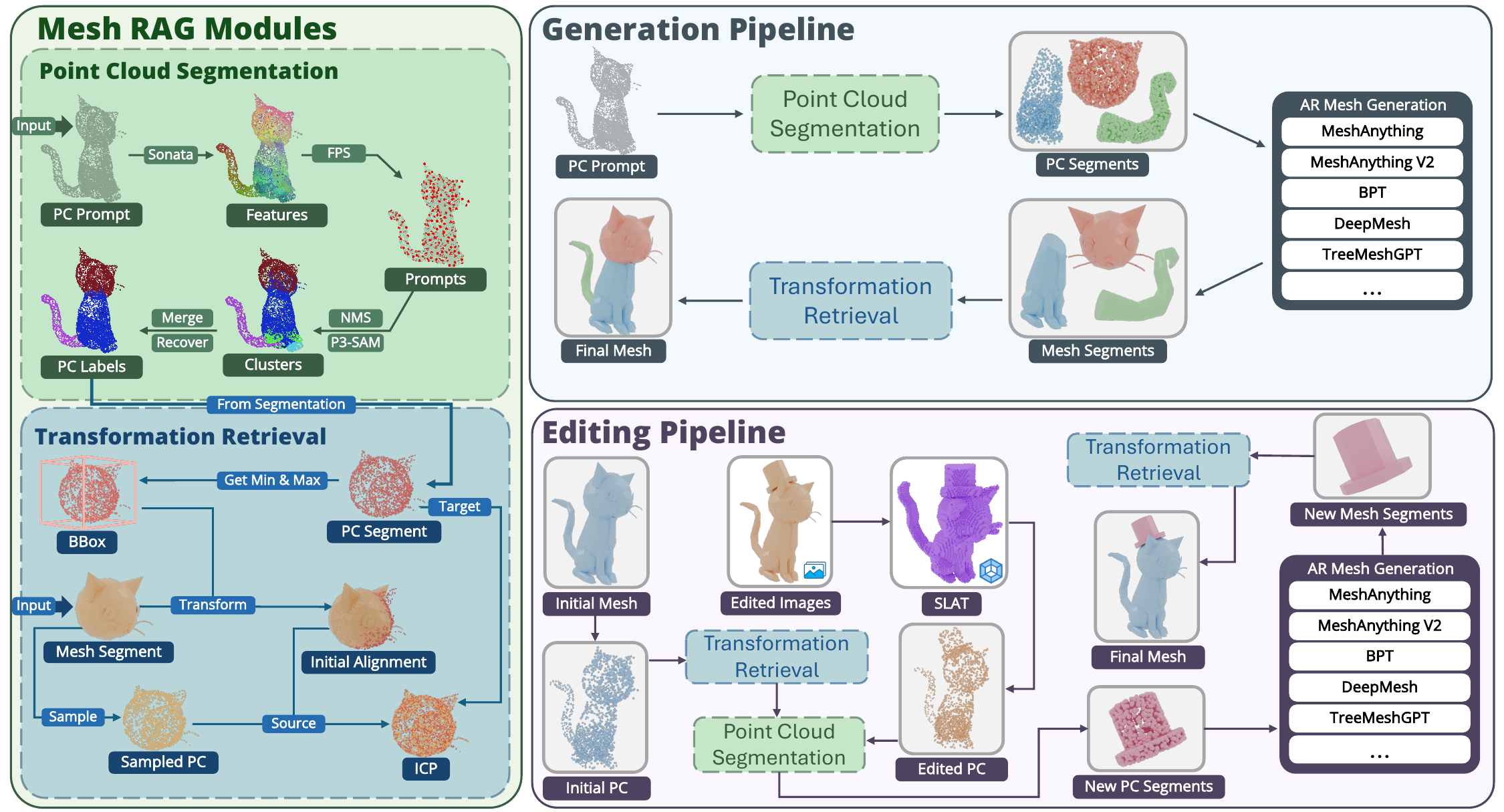}
    \caption{\textbf{Workflows for core components of Mesh RAG, parallel generation, and editing.} Mesh RAG includes a point cloud segmentation module that leverages Sonata \cite{wu2025sonata} and P3-SAM \cite{ma2025p3} for part segmentation on point clouds. It also includes a transformation retrieval module that first performs an initial coarse alignment by leveraging the known transformation for the point cloud prompt, and then using iterative closest point to refine the alignments.}
    \label{fig:workflows}
    \vspace{-0.4cm}
\end{figure*}

\section{Related Works}
\label{sec:related_works}

\subsection{3D Mesh Generation}

Research in 3D generation aims to automate mesh production for downstream applications like game development, simulation, and robotics \cite{hedden2025virtual, qiao2023drive, sun2022multi}. The generation of 3D content has seen rapid progress, though much initial work focused on representations like Neural Radiance Fields (NeRF) \cite{poole2022dreamfusion, mildenhall2021nerf, ni2025homer} and Gaussian Splatting \cite{lin2023magic3d, kerbl20233d}. While powerful for novel view synthesis, these representations are often not compatible with established graphics pipelines \cite{chen2023mobilenerf, li2024mixrt}. Triangular meshes remain the industry standard \cite{kobbelt2004survey}, but their irregular topology presents significant challenges for neural tokenization and embedding \cite{vaska2025reducing, chenmeshanything}.

Recent research in direct mesh generation has bifurcated into two main categories: reconstruction-based and autoregressive methods \cite{liu2024comprehensive}.
Reconstruction-based methods first generate an intermediate 3D representation and then apply a reconstruction algorithm to extract the final mesh. For instance, MeshLRM predicts a triplane-based NeRF and extracts the mesh via Differentiable Marching Cubes \cite{wei2024meshlrm, wei2025neumanifold}. SuGaR applies Poisson reconstruction to a generated Gaussian Splatting field \cite{guedon2024sugar, kazhdan2006poisson}, and TRELLIS decodes a mesh from a sparse 3D voxel grid \cite{xiang2025structured}. A common limitation of these approaches is topology quality. The extracted mesh topology is often dense, dictated by the reconstruction algorithm's parameters (e.g., voxel resolution) rather than the object's geometric features \cite{chenmeshanything, newman2006survey}. This issue results in dense meshes that are inefficient for real-time applications and difficult for artists to edit or animate \cite{liao2018deep}.

\subsection{Autoregressive Models and RAG}

Leveraging the transformer architecture \cite{vaswani2017attention} and scaling laws \cite{kaplan2020scaling}, autoregressive models have achieved great success in language modeling \cite{brown2020language, achiam2023gpt} and 2D image generation \cite{tian2024visual, yu2025randomized}. Inspired by this paradigm, researchers began leveraging transformers for mesh generation to produce artistic topologies \cite{chenmeshanything}. PolyGen was a pioneering work in this direction, directly outputting shape tokens \cite{nash2020polygen}. Subsequent research has largely focused on improving tokenization and compression, or adapting LLM techniques. For example, MeshAnything enables point cloud conditioned generation \cite{chenmeshanything}, and its successor MeshAnything V2 improves tokenization efficiency \cite{chen2025meshanything}. BPT and TreeMeshGPT further improve compression by exploiting geometric structure \cite{weng2025scaling, lionar2025treemeshgpt}. Others have adapted techniques like RLHF, with DeepMesh applying DPO after pre-training \cite{zhao2025deepmesh, ouyang2022training, rafailov2023direct}. Despite these advances, autoregressive mesh models face critical bottlenecks from their sequential paradigm. Their token-heavy nature creates a severe quality-efficiency trade-off, leading to slow inference, strong sequential dependency, and significant challenges for parallelization and incremental editing. 

To address similar problems in language modeling, Retrieval-Augmented Generation (RAG) retrieves relevant information and incorporates it into generation \cite{lewis2020retrieval}. This approach can improve speed, decouple sequential dependency, and enable longer effective context \cite{asai2024self, li2024long, ou2025accelerating}. Its core idea has also been extended to other domains; for example, AR-RAG adapts it for image generation, using K-Nearest Neighbor (KNN) for retrieving relevant patches instead of standard similarity metrics, which achieves notable improvements in accuracy and fidelity \cite{qi2025ar}.

However, this paradigm has not been applied to autoregressive mesh generation. Prior work focuses on optimizing the model (e.g., tokenization, training) rather than the inference process. We bridge this gap with Mesh RAG, a novel, training-free framework that re-conceptualizes generation as a compositional task. By leveraging retrieval of 3D transformation, we generate and integrate independent mesh components in parallel, directly addressing the core limitations of sequential dependency, speed, and editability without model retraining.

\section{Method}
\label{sec:method}

% \subsection{Overview}
In this work, we propose the Mesh RAG framework, which is designed to boost the capabilities of existing autoregressive mesh generation models. As Figure \ref{fig:workflows} illustrates, by leveraging the core components of Mesh RAG (point cloud segmentation and transformation retrieval), we enable these models to perform two novel abilities: parallel generation, which improves generation quality, and incremental editing, which makes the autoregressive paradigm more flexible.

% \subsection{Mesh RAG}
We focus on autoregressive mesh generation models that take a point cloud $\mathbf{P} \subset \mathbb{R}^3$ as input and output a token sequence, which is then detokenized into a triangular mesh $\mathbf{M}$. This formulation aligns with the predominant input modality for state-of-the-art autoregressive mesh generators \cite{chenmeshanything, chen2025meshanything, lionar2025treemeshgpt, zhao2025deepmesh, weng2025scaling, tangedgerunner, hao2024meshtron}.

Similar to RAG for other modalities \cite{lewis2020retrieval, radeva2024similarity, qi2025ar}, Mesh RAG enhances a generator with an external, non-parametric knowledge base. To apply this concept, we reframe the entire generation process. We first employ a point cloud segmentation module that serves an indexing purpose, decomposing the input $\mathbf{P}$ into a set of distinct parts, $\{\mathbf{P}_i\}$. The autoregressive model then generates a normalized mesh part $\mathbf{M}_i$ for each $\mathbf{P}_i$. This stage is where the retrieval-augmentation occurs: our transformation retrieval module then retrieves the original spatial context (the scale and pose) for each generated $\mathbf{M}_i$ by using its corresponding $\mathbf{P}_i$ as the key. This retrieved transformation augments the normalized mesh part, placing it correctly in the final, assembled object. While Mesh RAG operates on this point cloud representation, it remains compatible with other input modalities (e.g., text or image), as we elaborate in the supplementary material \S\ref{sec:supp_other_input_modalities}.

\subsection{Point Cloud Segmentation}
To address the bottlenecks of sequential generation (i.e., slow inference, long-range dependency, and limited context), we adopt a divide-and-conquer strategy. This strategy serves as the indexing step for our RAG pipeline, which is based on the insight that complex objects are compositional, and their constituent parts can be treated as individual meshes \cite{yang2025holopart}. This part-based approach has been successfully leveraged by recent reconstruction-based models \cite{liu2024part123, lin2025partcrafter}. However, existing part-based methods are designed for reconstruction-based models and typically require full training or fine-tuning. This makes them incompatible with pre-trained autoregressive models and unsuitable for a universal, plug-and-play framework. We, therefore, build upon the divide-and-conquer insight but apply it in a novel, training-free manner.

To perform this indexing via segmentation, we employ P3-SAM, a state-of-the-art 3D segmentation model \cite{ma2025p3}. P3-SAM uses a Sonata feature extractor \cite{wu2025sonata} and multi-scale segmentation heads. This architecture allows for interactive segmentation, such as isolating a specific region by clicking on a point, similar to Segment Anything (SAM) \cite{kirillov2023segment}. For our automated framework, we instead repurpose the original automatic segmentation pipeline from P3-SAM, which was designed for meshes, adapting it to work with point clouds. As illustrated in the upper left section of Figure \ref{fig:workflows}, this adapted pipeline first extracts features with Sonata, then uses Farthest Point Sampling (FPS) to generate prompts \cite{li2024deep}. It clusters the resulting masks using Non-Maximum Suppression (NMS) \cite{girshick2014rich, szeliski2022computer}, recovers unmasked regions with high spatial overlap, and finally assigns a unique label to each cluster.
This process takes the full point cloud $\mathbf{P}$ (with $N_{point}$ points) and produces a mask $\mathbf{K}_{point} \in \{0, 1, \ldots, N_{part}\}^{N_{point}}$, assigning each point to one of $N_{part}$ segments. Full implementation details are in the supplementary material \S\ref{sec:supp_seg_algo}. Using $\mathbf{K}_{point}$, we decompose $\mathbf{P}$ into $N_{part}$ point cloud segments $\{\mathbf{P}_i\}_{i=1}^{N_{part}}$. These segments are the keys our framework will use for both generation and retrieval.

A critical challenge arises here. Due to the normalization (e.g., centering and scaling) used in the training pipelines of most autoregressive models, if we feed each segment $\mathbf{P}_i$ directly into the model, the resulting mesh segments $\mathbf{M}_i$ will lose its transformation information relative to the original, complete mesh $\mathbf{M}$. Therefore, our RAG framework is designed to let the generator produce the normalized shape $\mathbf{M}_i$, while the spatial transformation is retrieved as the non-parametric context, as detailed next.

\subsection{Transformation Retrieval}

The core idea of RAG is to enhance generation with contextual, non-parametric information retrieved at runtime. These retrieval methods are modality-specific: RAG for LLMs uses similarity metrics to find relevant text \cite{lewis2020retrieval, radeva2024similarity}, while RAG for image generation uses patch-level KNN to find visual references \cite{qi2025ar}.

For 3D mesh generation, the crucial non-parametric information is the spatial context (position, rotation, and scale) of each segment. Our retrieval method is therefore a two-stage process to recover the correct affine transformation for each generated mesh segment $\mathbf{M}_i$, as depicted in the bottom left section of Figure \ref{fig:workflows}.

The first stage is a coarse alignment that computes an initial transformation $\mathbf{T}_{\text{restore}, i} \in SE(3)$ by aligning the normalized, generated mesh $\mathbf{M}_i$ to its corresponding original point cloud segment $\mathbf{P}_i$. This is achieved by matching the Axis-Aligned Bounding Boxes (AABB) of the two geometries. First, we compute the AABB for $\mathbf{M}_i$, yielding its center $\mathbf{c}_{M_i} \in \mathbb{R}^3$ and its extents (dimensions) $\mathbf{e}_{M_i} = (w_M, h_M, d_M)$. Second, we compute the AABB for $\mathbf{P}_i$, which gives its center $\mathbf{c}_{P_i} \in \mathbb{R}^3$ and its dimensional extents $\mathbf{e}_{P_i} = (w_P, h_P, d_P)$.

We then derive a non-uniform scaling vector $\mathbf{s}_i = (w_P/w_M, h_P/h_M, d_P/d_M)$. Using homogeneous coordinates, we define our $4 \times 4$ affine transformation as the product of three matrices. Let $\mathbf{T}(\mathbf{v}) \in SE(3)$ be the $4 \times 4$ translation matrix corresponding to a vector $\mathbf{v} \in \mathbb{R}^3$, $\mathbf{S}(\mathbf{s}) \in SE(3)$ be the $4 \times 4$ scaling matrix defined by the scaling vector $\mathbf{s} \in \mathbb{R}^3$, and let $\mathbf{T}(-\mathbf{c}_{M_i}) \in SE(3)$ be the $4 \times 4$ translation matrix corresponding to a vector $-\mathbf{c}_{M_i} \in \mathbb{R}^3$. The complete coarse transformation matrix $\mathbf{T}_{\text{restore}, i}$ is the product of these matrices, which first translates the mesh to the origin, scales it, and then translates it to the target position:
\begin{equation}
\label{eq:coarse}
\mathbf{T}_{\text{restore}, i} = \mathbf{T}(\mathbf{c}_{P_i}) \cdot \mathbf{S}(\mathbf{s}_i) \cdot \mathbf{T}(-\mathbf{c}_{M_i})
\end{equation}

We intentionally do not account for rotation in this initial alignment step. We observe that for these autoregressive mesh generation models, the orientation of the generated mesh $\mathbf{M}_i$ is usually already very close to that of its point cloud prompt $\mathbf{P}_i$. While small orientation deviations do exist—as the point cloud is only a prompt—we find these deviations are typically small enough to be reliably corrected by the subsequent ICP refinement stage.

In the second stage, we refine the alignment using Iterative Closest Point (ICP), a foundational point cloud registration method that has been widely applied to robotic localization \cite{zhang2021dspoint, sun2023benchmark, wu2024deep}. We first sample a new, dense point cloud $\mathbf{P}'_i$, including normal vectors, from the surface of $\mathbf{M}_i$. We apply the coarse transformation $\mathbf{T}_{\text{restore}, i}$ to this new cloud, yielding a pre-aligned source cloud $\mathbf{P}''_{i}$. This cloud is then registered against the original point cloud segment $\mathbf{P}_i$, which serves as the target. Given that both our source and target clouds contain rich normal information, we employ a Point-to-Plane ICP variant for its robustness and accuracy \cite{chen1992object}. This variant minimizes the sum of squared distances from each source point to the tangent plane of its corresponding target point. Formally, the algorithm finds the residual rotation $\mathbf{R}_{\text{ICP}} \in SO(3)$ and translation $\mathbf{t}_{\text{ICP}}  \in \mathbb{R}^3$ that minimize the objective function $E$:
$$
E(\mathbf{R}_{\text{ICP}}, \mathbf{t}_{\text{ICP}}) = \sum_{(\mathbf{s}, \mathbf{t}) \in \mathcal{Q}} ((\mathbf{R}_{\text{ICP}}\mathbf{s} + \mathbf{t}_{\text{ICP}} - \mathbf{t}) \cdot \mathbf{n}_t)^2
$$
where $\mathcal{Q}$ is the set of corresponding point pairs, $\mathbf{s} \in \mathbf{P}''_{i}$ is a source point, $\mathbf{t} \in \mathbf{P}_i$ is its corresponding target point, and $\mathbf{n}_t$ is the normal vector at $\mathbf{t}$. Given the strong initialization, we initialize the ICP with an identity transformation and apply a standard refinement process.

This ICP step yields a small, residual transformation $\mathbf{T}_{\text{ICP}, i} \in SE(3)$. The final, optimized transformation $\mathbf{T}_{\text{final}, i} \in SE(3)$ applied to the original mesh segment $\mathbf{M}_i$ is the concatenation of both transforms:
$$
\mathbf{T}_{\text{final}, i} = \mathbf{T}_{\text{ICP}, i} \cdot \mathbf{T}_{\text{restore}, i}
$$
Applying this final transformation matrix to all vertices ensures that a generated mesh segments $\mathbf{M}_i$ can be precisely localized and transformed within a coherent final mesh $\mathbf{M}$.

\subsection{Parallel Generation}
\label{sec:parallel_generation}

By decomposing the input, Mesh RAG unlocks parallel inference for autoregressive mesh generation models. This is possible because modern generative transformers, while autoregressive in their sequence dimension, are trained with a batch dimension. We exploit this by feeding multiple, independent segments as a single batch, allowing the model to process them concurrently. In this way, we can improve the generation quality for meshes since each part can consume longer and independent context, while also reducing the inference time compared to generating each part sequentially.

As illustrated in the top right section of Figure \ref{fig:workflows}, the parallel generation pipeline first segments the input point cloud $\mathbf{P}$ into its constituent parts, $\{\mathbf{P}_i\}_{i=1}^{N_{\text{part}}}$. These segments are then processed in batches by the off-the-shelf autoregressive model to generate the corresponding mesh parts, $\{\mathbf{M}_i\}_{i=1}^{N_{\text{part}}}$. Once all segments are generated, our transformation retrieval module is applied to each $\mathbf{M}_i$ to find its final transformation $\mathbf{T}_{\text{final}, i}$. Applying these transformations restores each part to its correct pose and scale, assembling them into the final, unified mesh $\mathbf{M}$.

\subsection{Incremental Editing}
\label{sec:incremental_editing}

A key advantage of our framework is the ability to perform incremental editing, a capability not supported by previous autoregressive mesh generation methods. The bottom right section of Figure \ref{fig:workflows} shows an example workflow for this process. We assume the user starts with an initial mesh $\mathbf{M}_{\text{init}}$, which we can sample a corresponding point cloud $\mathbf{P}_{\text{init}}$, and provides one or more images $\{\mathbf{I}_k\}_{k=1}^{N_{image}}$ guiding the desired edit. We then leverage TRELLIS \cite{xiang2025structured} to convert the images $\{\mathbf{I}_k\}$ into a Structured Latent (SLAT) representation, which is fundamentally a voxel-based representation that we can sample the edited point cloud, $\mathbf{P}_{\text{edited}}$.

To perform the edit and isolate the new geometry, we first align the initial mesh $\mathbf{M}_{\text{init}}$ to the new edited point cloud $\mathbf{P}_{\text{edited}}$. We accomplish this by sampling a point cloud from $\mathbf{M}_{\text{init}}$ and using our transformation retrieval module to register it against $\mathbf{P}_{\text{edited}}$, finding the optimal alignment. Once $\mathbf{M}_{\text{init}}$ is aligned, we use it as a 3D mask to subtract the existing geometry. The remaining points form the residual point cloud $\mathbf{P}_{\text{res}}$, which represents only the newly added or modified segments. These new segments $\mathbf{P}_{\text{res}}$ are then passed to the autoregressive model to generate the new mesh parts $\mathbf{M}_{\text{res}}$. We retrieve their final transformations $\mathbf{T}_{\text{final}, \text{res}}$ and apply them to $\mathbf{M}_{\text{res}}$. The final edited mesh, $\mathbf{M}_{\text{edited}}$, is the combination of the remaining parts of $\mathbf{M}_{\text{init}}$ and the newly generated, transformed segments.

\begin{figure*}[t]
    \centering
    \includegraphics[width=\textwidth]{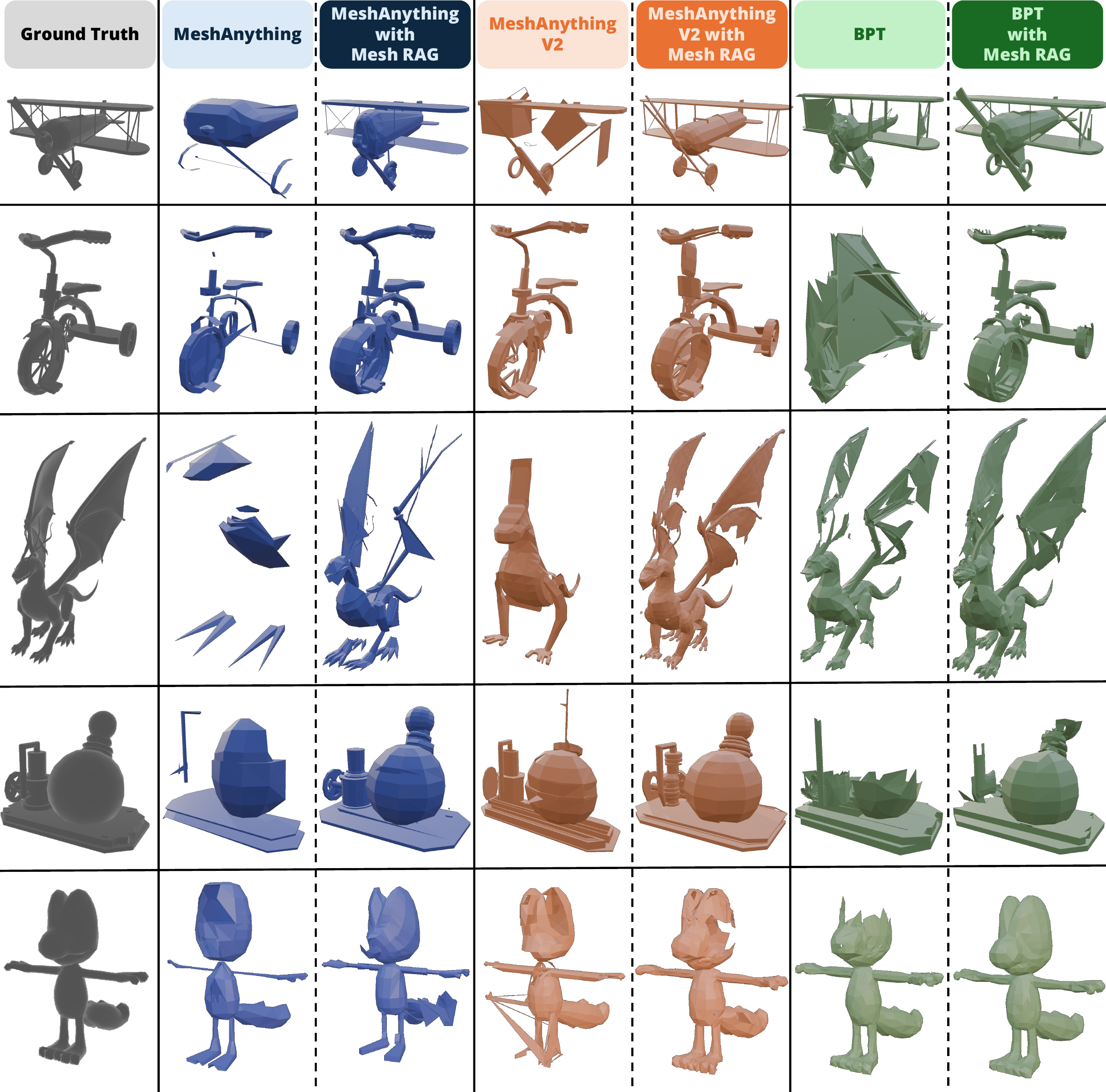}
    \caption{\textbf{Qualitative comparison of autoregressive mesh generation models with and without Mesh RAG.} Our plug-and-play framework significantly boosts the generation quality of baseline models (MeshAnything, MeshAnything V2, and BPT). By processing each segment with a dedicated context, Mesh RAG produces more complete geometries and preserves finer details. Additional qualtitative results for TreeMeshGPT and DeepMesh are in the supplementary material \S\ref{sec:supp_additional_qualtitative_results}.}
    \label{fig:qual_res}
\vspace{-0.2cm}
\end{figure*}

\section{Experiments}
\label{sec:experiments}

\subsection{Implementation Details}
We conduct all experiments on a workstation equipped with an AMD PRO 5975WX CPU, 128GB of RAM, and two NVIDIA RTX 4090 GPUs. To ensure fair evaluations, all pretrained models are used with their original, published parameters. This includes the segmentation models (Sonata, P3-SAM) \cite{wu2025sonata, ma2025p3}, the autoregressive generation models \cite{chenmeshanything, chen2025meshanything, weng2025scaling, zhao2025deepmesh, lionar2025treemeshgpt}, and the model used for the editing workflow \cite{xiang2025structured}. We refer readers to the original papers for specific hyperparameters, such as context length and point sampling configurations.

\subsection{Parallel Generation}

\paragraph*{\textbf{Procedure. }}

We quantitatively evaluate Mesh RAG on five baseline autoregressive models: MeshAnything \cite{chenmeshanything}, MeshAnything V2 \cite{chen2025meshanything}, Blocked and Patchified Tokenization (BPT) \cite{weng2025scaling}, TreeMeshGPT \cite{lionar2025treemeshgpt}, and DeepMesh \cite{zhao2025deepmesh}. For all Mesh RAG experiments, we use a consistent batch size of 8 for parallel generation. The evaluation is performed on 100 objects sampled from each of four datasets: ShapeNet \cite{chang2015shapenet}, Objaverse \cite{deitke2023objaverse}, Toys4k \cite{stojanov2021using}, and Thingi10k \cite{zhou2016thingi10k}. Table \ref{tab:summary_results} summarizes the averaged results, with full per-dataset breakdowns in \S\ref{sec:supp_comprehensive_quantitative_results}.

\paragraph*{\textbf{Results. }}

As shown in Table \ref{tab:summary_results}, applying Mesh RAG yields consistent and significant improvements across almost all fidelity metrics for every model tested. The performance lift is particularly striking for earlier models; for instance, MeshAnything and MeshAnything V2, when augmented with our framework, achieve geometric fidelity (e.g., $\text{CD}_{L1}$, HD, NC) that is competitive with, or even superior to, much larger, more recent models like BPT and DeepMesh in their baseline form.

Furthermore, our parallel processing approach accelerates inference for most of the larger models (MeshAnything V2, BPT, and DeepMesh). This demonstrates that the gains from batching segment generation outweigh the overhead of our framework's segmentation and retrieval stages. We note a minor increase in generation time for MeshAnything and TreeMeshGPT. This overhead can potentially be further alleviated by leveraging more powerful compute to enable higher batch sizes, as analyzed in our ablation study (\S\ref{sec:abl_study}). But in all cases, this slight time trade-off is met with a dramatic improvement in all geometric fidelity and completeness metrics.

Figure \ref{fig:qual_res} provides a qualitative comparison of several foundational autoregressive models, both with and without our Mesh RAG framework. Corroborating our quantitative findings, the effectiveness of Mesh RAG is particularly noticeable when applied to models with limited context lengths. Additional qualitative results with TreeMeshGPT and DeepMesh are provided in \S\ref{sec:supp_additional_qualtitative_results}.

\begin{table*}[t]
\caption{\textbf{Quantitative comparison of parallel generation performance.} We evaluate five autoregressive baselines—MeshAnything (MA), MeshAnything V2 (MA2), Blocked and Patchified Tokenization (BPT), TreeMeshGPT (TM-GPT), and DeepMesh (DM)—with and without our Mesh RAG (M-RAG) framework. Reported metrics include model parameters (P), generation time in minutes (T), L1 and L2 Chamfer Distance ($\text{CD}_{L1}$, $\text{CD}_{L2}$), Hausdorff Distance (HD), Normal Consistency (NC), F-Score (F1), Edged Chamfer Distance (ECD), and Edged F-Score (EF1). $\downarrow$ indicates lower is better, while $\uparrow$ indicates higher is better. The best and second best score are highlighted in \best{bold with underline} and \secondbest{bold only}. Per-dataset results and explanation for metrics are in the supplementary material (\S\ref{sec:supp_comprehensive_quantitative_results} and \S\ref{sec:supp_evaluation_metrics}).}
\centering
\label{tab:summary_results}
\begin{tabular}{lccccccccc}
\toprule
\textbf{Method} & \textbf{P} & \textbf{T (min.) $\downarrow$} & \textbf{$\text{CD}_{L1} \downarrow$} & \textbf{$\text{CD}_{L2} \downarrow$} & \textbf{HD $\downarrow$} & \textbf{NC $\uparrow$} & \textbf{F1 $\uparrow$} & \textbf{ECD $\downarrow$} & \textbf{EF1 $\uparrow$} \\
\midrule
MA & \multirow{2}{*}{595.84} & \best{0.572} & 0.094 & 0.094 & 0.145 & 0.735 & 0.963 & 0.249 & 0.544 \\
MA w/ M-RAG &             & 1.642 & 0.066 & 0.061 & 0.110 & 0.787 & 0.992 & 0.195 & 0.652 \\
\cdashline{1-10}
MA2 & \multirow{2}{*}{500.66} & 0.967 & 0.072 & 0.067 & 0.107 & 0.777 & 0.978 & 0.184 & 0.671 \\
MA2 w/ M-RAG &             & \secondbest{0.618} & 0.054 & 0.049 & 0.089 & 0.804 & \best{0.996} & 0.154 & 0.745 \\
\cdashline{1-10}
BPT & \multirow{2}{*}{711.45} & 2.652 & 0.057 & 0.055 & 0.100 & 0.811 & 0.988 & 0.131 & 0.800 \\
BPT w/ M-RAG &             & 1.592 & \best{0.045} & \best{0.039} & \secondbest{0.076} & \secondbest{0.839} & \secondbest{0.995} & \best{0.108} & \best{0.858} \\
\cdashline{1-10}
TM-GPT & \multirow{2}{*}{420.50} & 0.930 & 0.053 & 0.053 & 0.095 & 0.824 & 0.987 & 0.164 & 0.743 \\
TM-GPT w/ M-RAG &             & 0.959 & \secondbest{0.047} & \secondbest{0.042} & \best{0.068} & \best{0.867} & 0.990 & 0.162 & 0.767 \\
\cdashline{1-10}
DM & \multirow{2}{*}{908.15} & 4.561 & 0.065 & 0.061 & 0.100 & 0.801 & 0.988 & 0.143 & 0.768 \\
DM w/ M-RAG &             & 4.077 & 0.050 & 0.046 & 0.088 & 0.831 & 0.993 & \secondbest{0.119} & \secondbest{0.824} \\
\bottomrule
\end{tabular}
\end{table*}

\subsection{Ablation Study}
\label{sec:abl_study}

\paragraph*{\textbf{Procedure. }} To further analyze Mesh RAG, we perform ablation on various aspects of Mesh RAG using the 100 sampled objects from Toys4k. We explain our choice of using Toys4k for ablation and following experiments in the supplementary material \S\ref{sec:supp_subsequent_exp_ds}.

\begin{figure}
\centering
\includegraphics[width=\columnwidth]{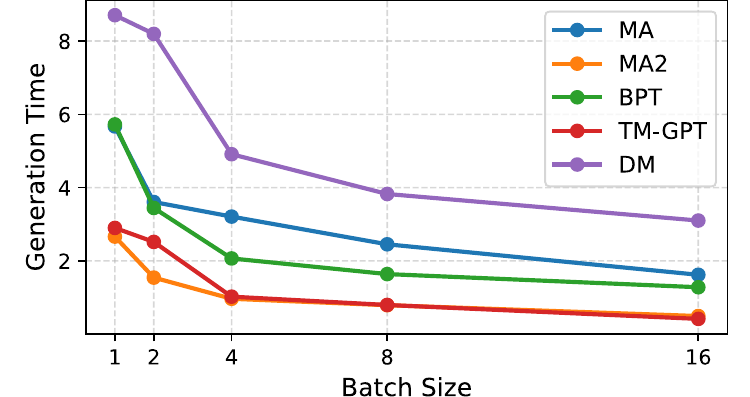}
\caption{\textbf{Generation time (minutes) as a function of batch size.} Each line corresponds to a different model augmented with our Mesh RAG framework.}
\label{fig:bs_gen_time}
\vspace{-15pt}
\end{figure}

\paragraph*{\textbf{Batch Size. }}
We first analyze the effect of batch size on the generation time of our parallel framework, with results shown in Figure \ref{fig:bs_gen_time}. For all five baseline models, increasing the batch size progressively reduces the total generation time compared to sequential part generation, demonstrating the throughput benefits of our parallel processing. This speedup is most pronounced for larger models; for instance, DeepMesh, the largest model in our comparison, sees the sharpest reduction in generation time when increasing the batch size from 1 to 4.

However, the performance gains show diminishing returns, and the speedup begins to saturate for all models at batch sizes larger than 4, which we hypothesize as a result of our specific hardware limitations. While larger batches still fit into VRAM, inference throughput becomes bounded by the available CUDA core compute \cite{recasens2025mind, lee2012cuda}. This suggests that scaling to more powerful compute hardware could further decrease generation times.

\paragraph*{\textbf{Effectiveness of Each Stage.}}
We also ablate each stage of the Mesh RAG framework to quantify its individual impact on generation time and mesh fidelity. For this test, we use MeshAnything V2, as it provides a strong balance between quality and speed based on our previous experiments.

The results in Table \ref{tab:ablation} break down the process. As expected, the generation step is the primary computational cost. Of our framework's components, segmentation is the most significant overhead, while the entire two-stage retrieval process is extremely fast at a combined 0.042s.

Crucially, the table highlights the value of our retrieval process. The raw, unaligned output from generation is geometrically poor. The coarse alignment step provides a massive fidelity boost, and the subsequent ICP refinement further polishes the result. This progressive alignment validates our divide-and-conquer approach, showing that the retrieval stages are both highly effective and efficient.

\begin{table}[h!]
    \centering
    \begin{tabular}{lcccc}
    \toprule
    \textbf{Stage} & \textbf{T (sec.) $\downarrow$} & \textbf{$\text{CD}_{L2} \downarrow$} & \textbf{NC $\uparrow$} & \textbf{F1 $\uparrow$} \\
    \midrule
    1 - PC Seg.  & 5.147 & -- & -- & -- \\
    2 - Mesh Gen.  & 42.280 & 0.260 & 0.575 & 0.772 \\
    3 - Coarse Align. & 0.001 & 0.071 & 0.774 & 0.969 \\
    4 - ICP Refine.   & 0.041 & 0.069 & 0.790 & 0.989 \\
    \bottomrule
    \end{tabular}
    \caption{\textbf{Ablation of Mesh RAG pipeline stages using MeshAnything V2.} The stages include Point Cloud Segmentation (PC Seg.), Mesh Generation (Mesh Gen.), Coarse Alignment (Coarse Align.), and ICP Refinement (ICP Refine.). We report the computation time in seconds (T) for each stage and the resulting mesh quality metrics after each stage. Metrics are chosen to evaluate geometry ($\text{CD}_{L2}$), local detail (NC), and completeness (F1).}
    \label{tab:ablation} 
\end{table}

\subsection{Incremental Editing}

\paragraph*{\textbf{Procedure. }}
We evaluate our incremental editing pipeline, which uses Mesh RAG with MeshAnything V2. To simulate a common editing task on the sampled Toys4k set, we randomly remove half of the segments from each object to create an initial mesh. We then use multi-view renderings (top, front, left) of the complete object as the guiding images, following the workflow in \S\ref{sec:incremental_editing}. As our framework is the first to enable incremental editing for autoregressive models, we compare against leading reconstruction-based editing methods: TRELLIS \cite{xiang2025structured} and Instant3dit \cite{barda2025instant3dit}. Following previous work on reconstruction-based editing, we use Peak Signal-to-Noise Ratio (PSNR), Structural Similarity Index Measure (SSIM), and Learned Perceptual Image Patch Similarity (LPIPS) as metrics for visual fidelity \cite{xiang2025structured, bovik2009essential, wang2004image, zhang2018unreasonable}.

\paragraph*{\textbf{Results. }}

\begin{figure}
\centering
\includegraphics[width=\columnwidth]{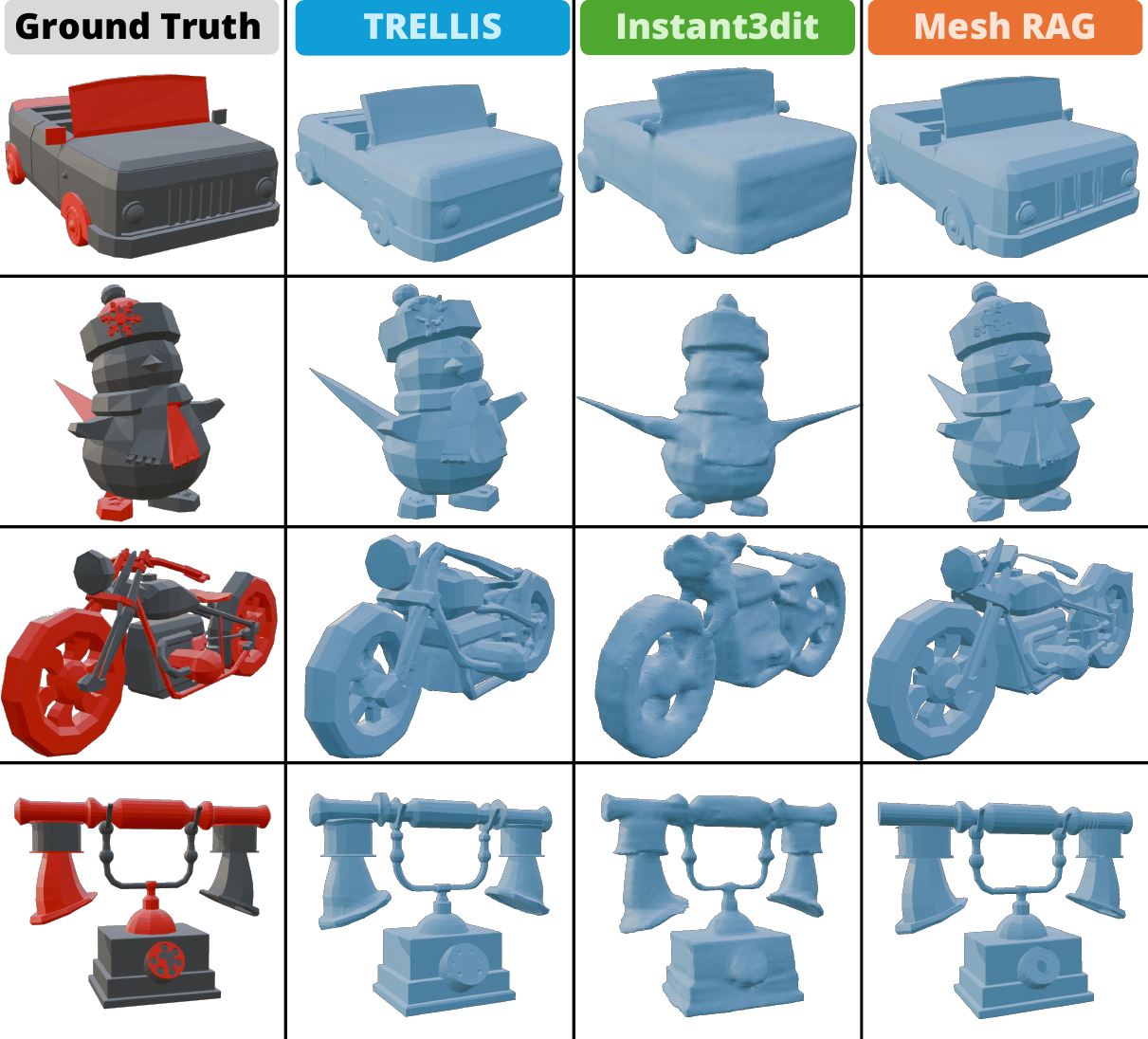}
\caption{\textbf{Qualitative comparison of incremental editing.} The Ground Truth visualizes the task: adding the red segments to the initial grey mesh. All generated outputs are rendered in a uniform cyan for fair comparison. }
\label{fig:edit_qualtitative}
\vspace{-10pt}
\end{figure}

As shown in Table \ref{tab:editing}, Mesh RAG achieves better fidelity across all rendering metrics. We attribute this to the autoregressive model's ability to generate high-frequency details, which are often lost during the Marching Cubes reconstruction in the other methods. Furthermore, our approach produces meshes with a substantially lower face count, demonstrating that it generates far more compact and artist-friendly topologies. Visual results in Figure \ref{fig:edit_qualtitative} corroborate the quantitative findings, showing that Mesh RAG can leverage autoregressive model to generate crisp local details than prior reconstruction methods.

\begin{comment}
\begin{table}[h!]
    \centering
    \begin{tabular}{lcccc}
    \toprule
    \textbf{Method} & \textbf{FC $\downarrow$} & \textbf{PSNR $\uparrow$} & \textbf{SSIM $\uparrow$} & \textbf{LPIPS $\downarrow$}  \\
    \midrule
    % TRELLIS: time for Slat generation: 2.54s; time for mesh decoding: 1.09s
    % Avg. time for TRELLIS: 3.638
    % Avg. time for Instant3dit: 10.027
    % Avg. time for Mesh RAG: 18.745
    TRELLIS      & \secondbest{11174}  & \secondbest{23.418}  & \secondbest{0.906} & \secondbest{0.109} \\  
    Instant3dit  & 59284  & 20.305  & 0.893 & 0.182 \\
    Mesh RAG     & \best{5384}   & \best{26.674} & \best{0.927} & \best{0.067} \\
    \bottomrule
    \end{tabular}
    \caption{\textbf{Quantitative comparison for incremental mesh editing.} We compare our method against reconstruction-based baselines. Our approach achieves superior rendering quality (PSNR, SSIM, LPIPS) and produces significantly more compact topology as indicated by Face Count (FC).}
    \label{tab:editing} 
\end{table}    
\end{comment}

\begin{table}[h!]
    \centering
    \small
    \setlength{\tabcolsep}{4pt}
    \begin{tabular}{l@{\hspace{4pt}}ccccc}
    \toprule
    \textbf{Method} & \textbf{FC $\downarrow$} & \textbf{PSNR $\uparrow$} 
    & \textbf{SSIM $\uparrow$} & \textbf{LPIPS $\downarrow$} & \textbf{T (sec.) $\downarrow$} \\
    \midrule
    TRELLIS      & \secondbest{11174} & \secondbest{23.418} & \secondbest{0.906} 
                 & \secondbest{0.109} & 3.64 \\
    Instant3dit  & 59284              & 20.305              & 0.893 
                 & 0.182              & 10.03 \\
    Mesh RAG     & \best{5384}        & \best{26.674}       & \best{0.927} 
                 & \best{0.067}       & 18.75 \\
    \bottomrule
    \end{tabular}
    \caption{\textbf{Quantitative comparison for incremental mesh editing.}  
    Our method outperforms baselines in reconstruction quality (PSNR, SSIM, LPIPS) 
    and yields more compact topology (Face Count, FC), though it takes more computation time (T).}
    \label{tab:editing}
    \vspace{-10pt}
\end{table}

\section{Discussion}
\label{sec:conclusion}
In this work, we introduced Mesh RAG, a training-free, plug-and-play framework that unlocks parallel generation and incremental editing for autoregressive mesh models. By employing a divide-and-conquer strategy, our method decouples the generation task from its inherent sequential dependency. Instead of generating one long, error-prone sequence, Mesh RAG segments the input and allows the baseline model to generate each part with a focused, standalone context. This approach mitigates error accumulation, significantly boosting fidelity, and enables parallel, batched inference to accelerate generation speed.

Despite its strong performance, Mesh RAG has clear limitations that open avenues for future work. First, the framework generates all segments independently, including semantically identical parts like chair legs or vehicle wheels. A future direction is to incorporate a component instancing mechanism to enforce geometric consistency and further reduce generation time. Second, our compatibility with text or image inputs currently relies on an intermediate SLAT representation. Developing a native, end-to-end multi-modal framework is another promising direction.

{
    \small
    \bibliographystyle{ieeenat_fullname}
    \bibliography{main}
}

 \clearpage
\setcounter{page}{1}
\maketitlesupplementary

% 
%Having the supplementary compiled together with the main paper means that:
% 
%\begin{itemize}
%\item The supplementary can back-reference sections of the main paper, for example, we can refer to \cref{sec:intro};
%\item The main paper can forward reference sub-sections within the supplementary explicitly (e.g. referring to a particular experiment); 
%\item When submitted to arXiv, the supplementary will already included at the end of the paper.
%\end{itemize}
% 
%To split the supplementary pages from the main paper, you can use \href{https://support.apple.com/en-ca/guide/preview/prvw11793/mac#:~:text=Delete%20a%20page%20from%20a,or%20choose%20Edit%20%3E%20Delete).}{Preview (on macOS)}, \href{https://www.adobe.com/acrobat/how-to/delete-pages-from-pdf.html#:~:text=Choose%20%E2%80%9CTools%E2%80%9D%20%3E%20%E2%80%9COrganize,or%20pages%20from%20the%20file.}{Adobe Acrobat} (on all OSs), as well as \href{https://superuser.com/questions/517986/is-it-possible-to-delete-some-pages-of-a-pdf-document}{command line tools}.

\section{Automatic Segmentation Algorithm}
\label{sec:supp_seg_algo}

Algorithm \ref{alg:point_cloud_seg} details our automatic segmentation process. Given an input point cloud $\mathbf{P} \in \mathbb{R}^{N_p \times 3}$ and normals $\mathbf{N}$, we first normalize the coordinates and extract geometric features $\mathbf{f}_p$ using the Sonata backbone \cite{wu2025sonata}. To discover potential parts without manual intervention, we sample $N_{prompt}$ point prompts $\{\mathbf{p}_j\}_{j=1}^{N_{prompt}}$ via Farthest Point Sampling (FPS) \cite{li2024deep}. For each prompt $\mathbf{p}_j$, the P3-SAM decoder predicts three candidate masks $\mathbf{k}_j^{(1)}, \mathbf{k}_j^{(2)}, \mathbf{k}_j^{(3)}$ with associated IoU scores $\mathbf{z}_j$; we retain the mask $\mathbf{k}_j$ corresponding to the maximum score $z_j$ \cite{ma2025p3}. The resulting set of candidate masks is sorted by score and filtered using Non-Maximum Suppression (NMS) with threshold $\tau_{NMS}$ to produce an initial set of clusters $\mathcal{C}$ \cite{girshick2014rich, szeliski2022computer}. We filter out negligible clusters and further merge spatially adjacent clusters based on IoU of oriented bounding boxes to form the refined set $\mathcal{C}'$. To ensure comprehensive coverage, we initialize a binary mask $\mathbf{u} \leftarrow \mathbf{1}_{N_p}$ to track unassigned points. We then iteratively revisit the discarded masks to recover those that are predominantly composed of unassigned points. Specifically, a mask is reinstated if its intersection with $\mathbf{u}$ exceeds a recovery threshold $\tau_{recover}$. Finally, we assign a unique label to each point in the output $\mathbf{k}_{point}$ by projecting the clusters in $\mathcal{C}'$ onto the point cloud, prioritizing those with larger surface areas.

\begin{algorithm}[t]
\caption{Point Cloud Automatic Segmentation}
\label{alg:point_cloud_seg}
\begin{algorithmic}[1]
\INPUT Point cloud $\mathbf{P} \in \mathbb{R}^{N_p \times 3}$, normals $\mathbf{N}$, number of point prompts $N_{prompt}$
\OUTPUT Point labels $\mathbf{k}_{point} \in \{0, 1, \ldots, N_{point}\}^{N_p}$

\STATE $\mathbf{P} \leftarrow \text{Normalize}(\mathbf{P})$
\STATE $\mathbf{f}_p \leftarrow \text{Sonata}(\mathbf{P}, \mathbf{N})$
\STATE $\{\mathbf{p}_j\}_{j=1}^{N_{prompt}} \leftarrow \text{FPS}(\mathbf{P}, N_{prompt})$

\FOR{$j = 1$ to $N_{prompt}$}
    \STATE $\mathbf{k}_j^{(1)}, \mathbf{k}_j^{(2)}, \mathbf{k}_j^{(3)}, \mathbf{z}_j \leftarrow \text{P}^3\text{-SAM}(\mathbf{f}_p, \mathbf{P}, \mathbf{p}_j)$
    \STATE $\mathbf{k}_j \leftarrow \mathbf{k}_j^{(\arg\max \mathbf{z}_j)}$, $z_j \leftarrow \max(\mathbf{z}_j)$
\ENDFOR
\STATE Sort $\{\mathbf{k}_j, z_j\}$ by $z_j$ in descending order

\STATE $\mathcal{C} \leftarrow \text{NMS}(\{\mathbf{k}_j\}, \tau_{NMS})$
\STATE $\mathcal{C} \leftarrow \{c \in \mathcal{C} : |\mathcal{C}[c]| > 2\}$ 
\STATE Merge clusters via OBB IoU to produce $\mathcal{C}'$ 

\STATE $\mathbf{u} \leftarrow \mathbf{1}_{N_p}$
\FORALL{$c \in \mathcal{C}'$}
    \STATE $\mathbf{u}[\mathbf{k}_c] \leftarrow 0$
\ENDFOR
\FOR{$i \notin \mathcal{C}'$}
    \IF{$|\mathbf{k}_i \cap \mathbf{u}| / |\mathbf{k}_i| > \tau_{recover}$}
        \STATE $\mathcal{C}'[i] \leftarrow [i]$
        \STATE $\mathbf{u}[\mathbf{k}_i] \leftarrow 0$
    \ENDIF
\ENDFOR

\STATE $\mathbf{k}_{point} \leftarrow \mathbf{0}_{N_p}$
\FORALL{$c \in \mathcal{C}'$ sorted by area (descending)}
    \STATE $\mathbf{k}_{point}[\mathbf{k}_c] \leftarrow c$
\ENDFOR
\RETURN $\mathbf{k}_{point}$

\end{algorithmic}
\end{algorithm}

\section{Comprehensive Quantitative Results}
\label{sec:supp_comprehensive_quantitative_results}

\begin{table*}[t]
\centering
\caption{\textbf{Detailed quantitative evaluation per dataset.} This table expands upon the summary results in the main text (Table \ref{tab:summary_results}) by breaking down the performance of five autoregressive baselines with and without our M-RAG framework across four distinct datasets: ShapeNet, Objaverse, Toys4k, and Thingi10k. Notation and metrics follow the main text.}
\label{tab:comprehensive_quant_results}
% For the \cdashline alternative, you would add: \usepackage{arydshln} in your preamble
\begin{tabular}{llcccccccc}
\toprule
\textbf{Dataset} & \textbf{Method} & \textbf{T $\downarrow$} & \textbf{$\text{CD}_{L1} \downarrow$} & \textbf{$\text{CD}_{L2} \downarrow$} & \textbf{HD $\downarrow$} & \textbf{NC $\uparrow$} & \textbf{F1 $\uparrow$} & \textbf{ECD $\downarrow$} & \textbf{EF1 $\uparrow$} \\
\midrule
\multirow{10}{*}{ShapeNet}
& MA          & 0.456 & 0.070 & 0.072 & 0.128 & 0.825 & 0.981 & 0.309 & 0.514 \\
& MA w/ M-RAG & \secondbest{0.450} & 0.055 & 0.049 & 0.086 & 0.863 & 0.997 & 0.216 & 0.653 \\
\cdashline{2-10}
& MA2          & 0.459 & 0.045 & 0.040 & 0.070 & 0.853 & 0.993 & 0.183 & 0.675 \\
& MA2 w/ M-RAG & \best{0.288} & \best{0.036} & \best{0.032} & \secondbest{0.057} & \best{0.891} & \secondbest{0.998} & 0.181 & 0.693 \\
\cdashline{2-10}
& BPT          & 2.295 & 0.049 & 0.048 & 0.093 & 0.859 & 0.992 & \secondbest{0.175} & 0.695 \\
& BPT w/ M-RAG & 1.739 & 0.042 & 0.036 & 0.064 & 0.872 & \best{0.999} & \best{0.148} & \best{0.759} \\
\cdashline{2-10}
& TM-GPT          & 0.744 & 0.046 & 0.047 & 0.089 & 0.873 & 0.985 & 0.221 & 0.618 \\
& TM-GPT w/ M-RAG & 0.692 & \secondbest{0.039} & \secondbest{0.033} & \best{0.056} & \secondbest{0.887} & \best{0.999} & 0.204 & 0.640 \\
\cdashline{2-10}
& DM          & 4.807 & 0.064 & 0.060 & 0.102 & 0.848 & 0.985 & 0.213 & 0.652 \\
& DM w/ M-RAG & 4.489 & 0.060 & 0.055 & 0.104 & 0.848 & 0.989 & 0.164 & \secondbest{0.729} \\
\midrule
\multirow{10}{*}{Objaverse}
& MA          & \secondbest{0.599} & 0.112 & 0.116 & 0.177 & 0.649 & 0.941 & 0.208 & 0.640 \\
& MA w/ M-RAG & 2.158 & 0.066 & 0.062 & 0.131 & 0.733 & 0.996 & 0.158 & 0.674 \\
\cdashline{2-10}
& MA2          & 1.104 & 0.068 & 0.066 & 0.119 & 0.693 & 0.989 & 0.125 & 0.810 \\
& MA2 w/ M-RAG & \best{0.573} & 0.049 & 0.047 & 0.109 & 0.748 & \secondbest{0.998} & 0.097 & 0.890 \\
\cdashline{2-10}
& BPT          & 3.676 & 0.073 & 0.071 & 0.130 & 0.708 & 0.985 & 0.102 & 0.867 \\
& BPT w/ M-RAG & 1.558 & 0.048 & 0.044 & 0.102 & 0.756 & \best{0.999} & 0.083 & \secondbest{0.929} \\
\cdashline{2-10}
& TM-GPT          & 0.751 & 0.060 & 0.062 & 0.117 & 0.750 & 0.987 & 0.122 & 0.814 \\
& TM-GPT w/ M-RAG & 0.834 & 0.057 & 0.051 & \secondbest{0.089} & \best{0.821} & 0.990 & 0.124 & 0.854 \\
\cdashline{2-10}
& DM          & 5.693 & \secondbest{0.047} & \secondbest{0.043} & 0.091 & 0.763 & \secondbest{0.998} & \secondbest{0.073} & 0.918 \\
& DM w/ M-RAG & 5.254 & \best{0.039} & \best{0.035} & \best{0.074} & \secondbest{0.782} & \best{0.999} & \best{0.057} & \best{0.950} \\
\midrule
\multirow{10}{*}{Toys4k}
& MA          & \best{0.663} & 0.128 & 0.121 & 0.163 & 0.710 & 0.940 & 0.283 & 0.431 \\
& MA w/ M-RAG & 2.453 & 0.085 & 0.076 & 0.116 & 0.802 & 0.978 & 0.234 & 0.604 \\
\cdashline{2-10}
& MA2          & 1.237 & 0.113 & 0.102 & 0.137 & 0.769 & 0.943 & 0.260 & 0.522 \\
& MA2 w/ M-RAG & \secondbest{0.791} & 0.078 & 0.069 & 0.106 & 0.790 & \secondbest{0.989} & 0.195 & 0.672 \\
\cdashline{2-10}
& BPT &          2.803 & 0.068 & 0.063 & 0.104 & 0.818 & 0.978 & \secondbest{0.120} & 0.817 \\
& BPT w/ M-RAG & 1.640 & \secondbest{0.057} & \secondbest{0.049} & \secondbest{0.079} & \secondbest{0.857} & 0.983 & \best{0.097} & \best{0.877} \\
\cdashline{2-10}
& TM-GPT          & 0.863 & 0.063 & 0.059 & 0.090 & 0.830 & 0.983 & 0.162 & 0.766 \\
& TM-GPT w/ M-RAG & 0.797 & \best{0.050} & \best{0.046} & \best{0.065} & \best{0.887} & 0.986 & 0.169 & 0.805 \\
\cdashline{2-10}
& DM          & 4.876 & 0.101 & 0.098 & 0.134 & 0.756 & 0.974 & 0.157 & 0.714 \\
& DM w/ M-RAG & 3.826 & 0.065 & 0.059 & 0.107 & 0.828 & \best{0.990} & 0.121 & \secondbest{0.827} \\
\midrule
\multirow{10}{*}{Thingi10k}
& MA          & \best{0.570} & 0.066 & 0.067 & 0.113 & 0.756 & 0.988 & 0.195 & 0.589 \\
& MA w/ M-RAG & 1.505 & 0.056 & 0.055 & 0.105 & 0.750 & 0.996 & 0.170 & 0.675 \\
\cdashline{2-10}
& MA2          & 1.067 & 0.060 & 0.060 & 0.101 & 0.794 & 0.988 & 0.169 & 0.677 \\
& MA2 w/ M-RAG & \secondbest{0.818} & 0.052 & 0.049 & 0.085 & 0.788 & \secondbest{0.997} & 0.144 & 0.726 \\
\cdashline{2-10}
& BPT          & 1.832 & 0.039 & 0.038 & 0.074 & 0.858 & 0.995 & \secondbest{0.126} & \secondbest{0.819} \\
& BPT w/ M-RAG & 1.431 & \best{0.032} & \best{0.028} & \best{0.058} & \secondbest{0.872} & \best{0.999} & \best{0.103} & \best{0.866} \\
\cdashline{2-10}
& TM-GPT          & 1.362 & 0.043 & 0.043 & 0.084 & 0.843 & 0.994 & 0.151 & 0.774 \\
& TM-GPT w/ M-RAG & 1.514 & 0.041 & 0.038 & \secondbest{0.061} & \best{0.873} & 0.985 & 0.152 & 0.770 \\
\cdashline{2-10}
& DM          & 2.867 & 0.047 & 0.044 & 0.074 & 0.838 & 0.993 & 0.129 & 0.787 \\
& DM w/ M-RAG & 2.739 & \secondbest{0.037} & \secondbest{0.036} & 0.066 & 0.865 & 0.993 & 0.132 & 0.789 \\
\bottomrule
\end{tabular}
\end{table*}

Table \ref{tab:comprehensive_quant_results} provides a comprehensive breakdown of quantitative performance across the ShapeNet, Objaverse, Toys4k, and Thingi10k datasets. Consistent with the aggregated results in the main text, applying Mesh RAG yields robust improvements in geometric fidelity across all domains. The per-dataset analysis reveals that our framework is particularly effective at improving the geometric fidelity; for instance, BPT augmented with Mesh RAG achieves the best generation results in terms of all quantitative metrics on the Thingi10k dataset, significantly outperforming the baseline BPT.

Regarding computational efficiency, the detailed breakdown validates that our parallel generation strategy offers substantial speedups for some foundation models. Notably, on all datasets, MeshAnything V2 with Mesh RAG not only improves the generation quality but also reduces inference time by approximately $23\%\text{--}48\%$ compared to its serial baseline. While DeepMesh remains the most computationally intensive model, our framework consistently improves its fidelity without imposing additional time overheads in most cases. These results confirm that the benefits of Mesh RAG are not limited to specific object categories but generalize effectively across varied topological complexities and data distributions.

\section{Additional Qualitative Results}
\label{sec:supp_additional_qualtitative_results}

\begin{figure*}[p]
    \centering
    \includegraphics[width=\textwidth]{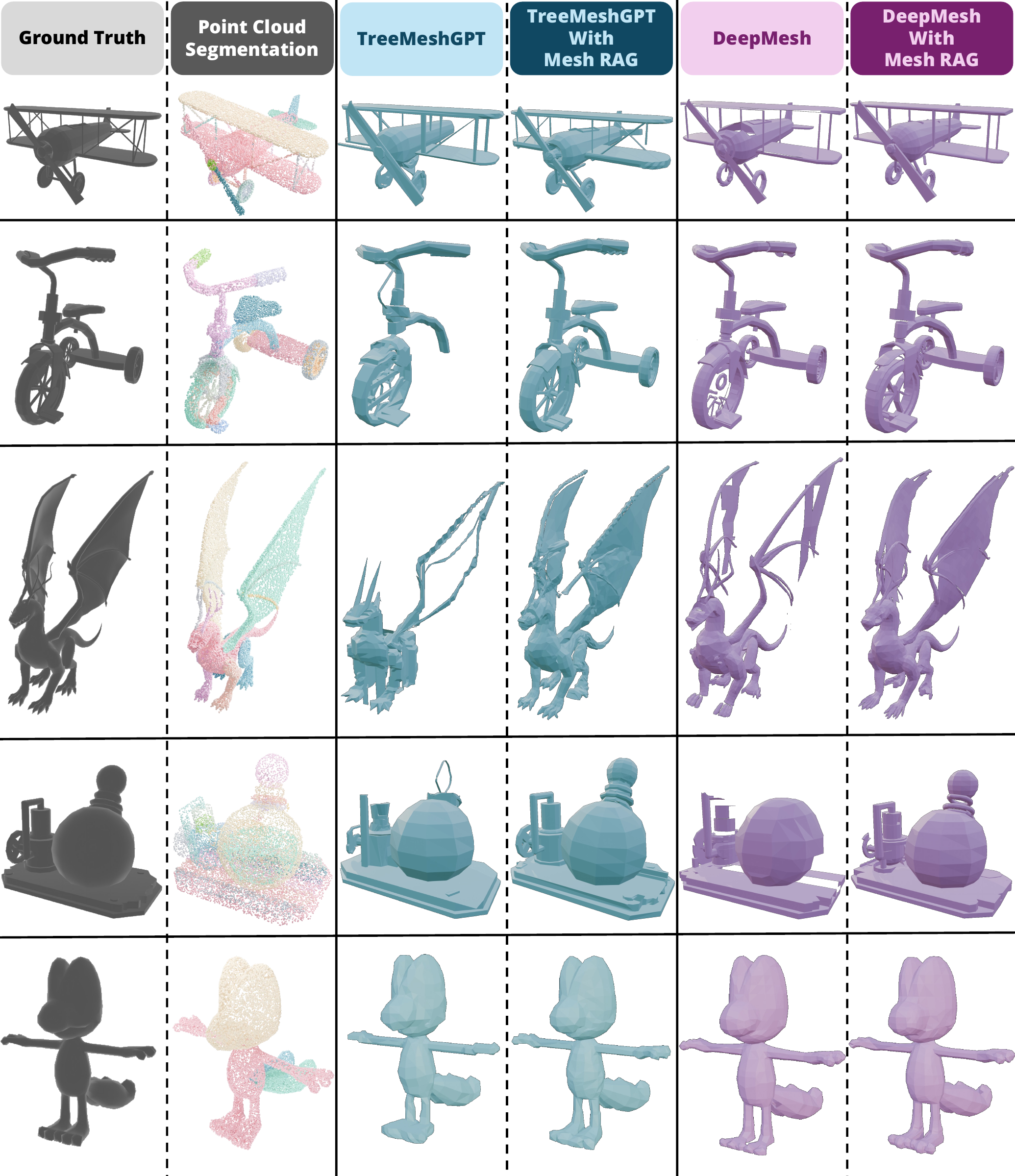}
    \caption{\textbf{Additional qualitative comparison for TreeMeshGPT and DeepMesh.} We present visual comparisons of generated meshes with the ground truth and the intermediate point cloud segmentation produced by our framework.}
    \label{fig:add_qual_res}
\end{figure*}

Figure \ref{fig:add_qual_res} shows our qualitative results of using Mesh RAG with TreeMeshGPT and DeepMesh. These results reinforce that baseline autoregressive models, particularly those with smaller capacities like TreeMeshGPT, often fail to generate complete geometry for complex objects. This is visibly manifested as wireframe-like sparsity (e.g., the dragon in the third row) or missing geometric connections or components (e.g., the tricycle wheel in the second row).

By explicitly conditioning the generation process on each segments, Mesh RAG ensures that each geometric component is attended to individually. Consequently, using Mesh RAG with TreeMeshGPT or DeepMesh recover thin structures and maintain global coherence that the baselines miss. This visual evidence aligns with our quantitative results, demonstrating that Mesh RAG allows baselines to produce dense, high-fidelity meshes.

\section{Evaluation Metrics}
\label{sec:supp_evaluation_metrics}

\begin{figure}[H]
\centering
\includegraphics[width=\columnwidth]{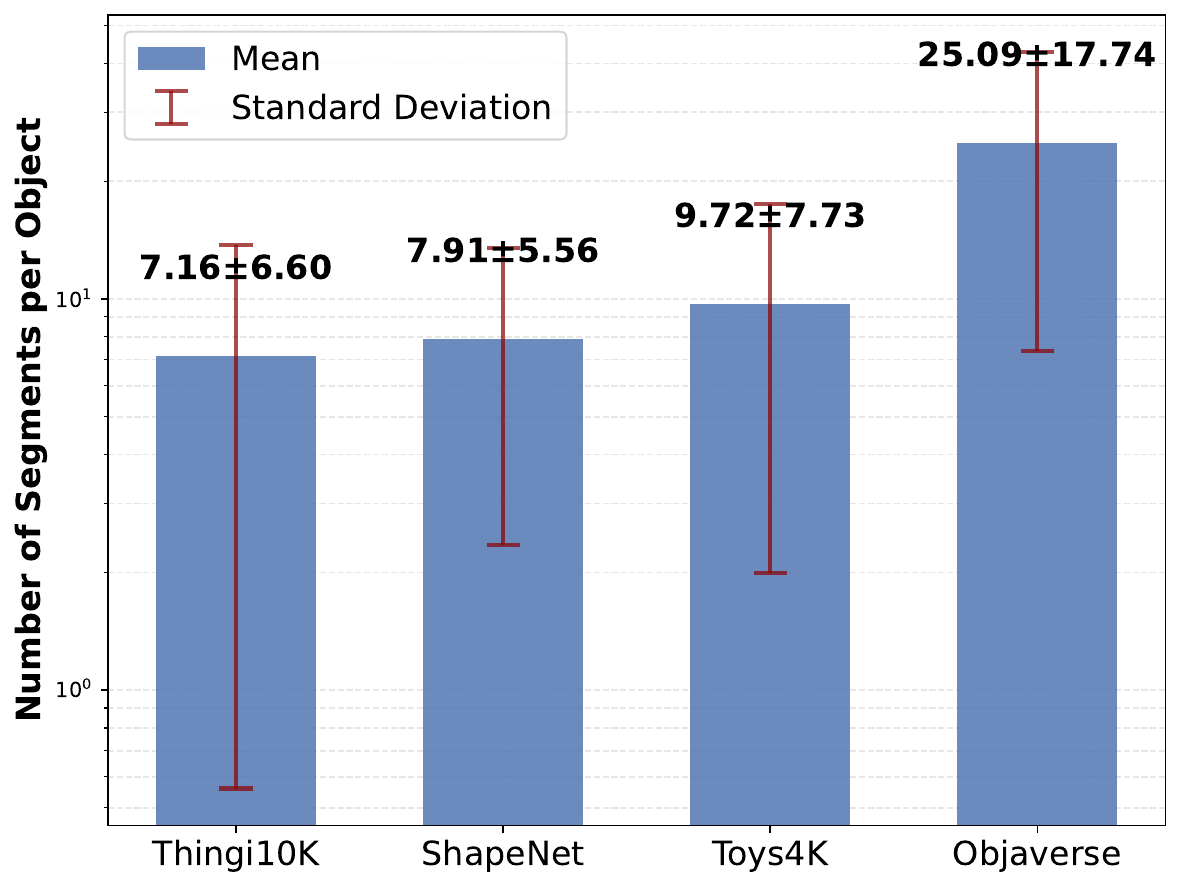}
\caption{\textbf{Segmentation statistics across evaluation datasets.} We plot the mean number of segments per object (log scale); error bars indicate standard deviation.}
\label{fig:metadata_eval_ds}
\end{figure}

To provide a comprehensive assessment of mesh quality, we employ a diverse set of geometric metrics. For all geometric metrics, we normalize both the generated mesh and the ground truth mesh.

\vspace{1mm}
\noindent\textbf{Chamfer Distance (CD).} We report both $\text{CD}_{L1}$ and $\text{CD}_{L2}$ to capture different aspects of geometric error \cite{fan2017point}. $\text{CD}_{L1}$ represents the mean of the minimum Euclidean distances between the sampled point sets, providing a robust measure of overall convergence. $\text{CD}_{L2}$, reported as the root-mean-square distance, penalizes large outliers more heavily, making it sensitive to severe local geometric distortions.

\vspace{1mm}
\noindent\textbf{Hausdorff Distance (HD).} To measure the worst-case geometric deviation, we compute the bidirectional Hausdorff Distance \cite{cignoni1998metro}. Unlike the averaging approach of Chamfer Distance, this metric identifies the single largest distance between any point in the predicted mesh and its nearest neighbor in the ground truth, effectively capturing the maximum local error.

\vspace{1mm}
\noindent\textbf{Normal Consistency (NC).} To evaluate the quality of surface details and high-frequency geometry, we compute Normal Consistency \cite{mescheder2019occupancy}. For each point in $\mathcal{P}$, we find the nearest neighbor in $\mathcal{G}$ and compute the absolute dot product of their normals, averaging this value bidirectionally. A higher score indicates better alignment of local surface orientation.

\vspace{1mm}
\noindent\textbf{F-Score (F1).} Following \cite{tatarchenko2019single}, we report the F-Score as the harmonic mean of precision and recall. Precision is the percentage of points in $\mathcal{P}$ within distance $\tau$ of any point in $\mathcal{G}$, and recall is the percentage of points in $\mathcal{G}$ within distance $\tau$ of any point in $\mathcal{P}$.

\vspace{1mm}
\noindent\textbf{Edge Chamfer Distance (ECD) and Edge F-Score (EF1).} Standard metrics often average out errors in sharp regions. To explicitly evaluate the preservation of sharp features, we employ edge-specific variants of CD and F1 \cite{chen2021neural}. We extract edge points from the sampled point clouds based on local normal variation.

\begin{figure}
\centering
\includegraphics[width=\columnwidth]{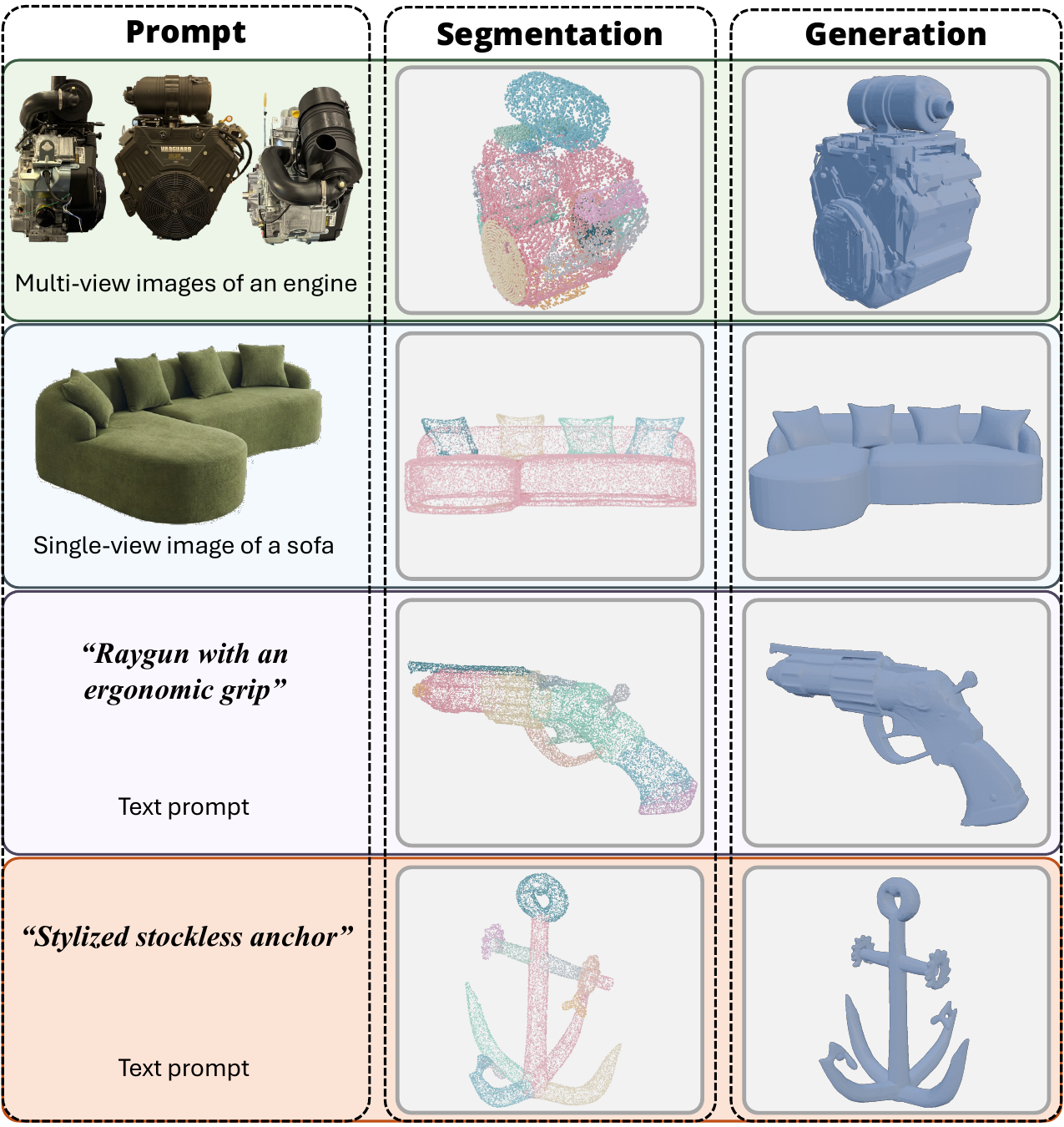}
\caption{\textbf{Generative results using image and text prompts.} We leverage TRELLIS to lift single-view images, multi-view inputs, or text descriptions into a point cloud representation via SLAT. Mesh RAG then performs parallel autoregressive generation on the segmented parts to produce the final mesh.}
\label{fig:other_modalities}
\end{figure}

\section{Evaluation Datasets}
\label{sec:supp_subsequent_exp_ds}

We conduct our ablation study and subsequent evaluations on the Toys4K dataset. We selected this dataset to ensure a representative evaluation of part-level generation without introducing bias. As detailed in Figure \ref{fig:metadata_eval_ds}, datasets like Thingi10K and ShapeNet are structurally simpler, averaging only $7.16$ and $7.91$ segments per object, respectively. This limited variance reduces the difficulty of the generation task. In contrast, Objaverse is an outlier with a high average segment count ($25.09 \pm 17.74$). Since our method's efficiency scales with the number of parts, using Objaverse would disproportionately favor our parallel architecture, potentially masking the specific contributions of the ablated components. Toys4K ($9.72 \pm 7.73$) offers a balanced complexity distribution, making it the most rigorous testbed.

\FloatBarrier
\newpage

\section{Generation with Other Input Modalities}
\label{sec:supp_other_input_modalities}

We demonstrate the versatility of Mesh RAG by extending it to image-to-mesh and text-to-mesh generation tasks. To bridge these modalities, we employ an intermediate 3D representation. Similar to the point cloud extraction process for editing described in \S\ref{sec:incremental_editing}, we utilize TRELLIS \cite{xiang2025structured} to generate a Structured Latent (SLAT) voxel representation from the input prompts. We then sample a point cloud from this voxel representation. By effectively treating the SLAT output of TRELLIS as a proxy, this pipeline allows Mesh RAG to inherit the multi-modal capabilities while enforcing autoregressive generation. Figure \ref{fig:other_modalities} illustrates successful generation examples across single-view images, multi-view images, and text prompts.

\end{document}